\documentclass[fleqn,10pt]{wlscirep}
\usepackage[utf8]{inputenc}
\usepackage[T1]{fontenc}
\usepackage{csquotes}
\usepackage{caption}
\usepackage{subcaption}
\usepackage{url} 
\usepackage{array}

\title{Video-based Exercise Classification and Activated Muscle Group Prediction with Hybrid X3D-SlowFast Network}

\author[1,*]{Manvik Pasula}
\author[2]{Pramit Saha}
\affil[1]{Independent Researcher, USA}
\affil[2]{University of Oxford, Department of Engineering Science, Oxford, OX1 4BH, United Kingdom}

\affil[*]{CORRESPONDING AUTHOR: Manvik Pasula (e-mail: manvikpasula@gmail.com)}


\begin{abstract}
This paper introduces a simple yet effective strategy for exercise classification and muscle group activation prediction (MGAP). These tasks have significant implications for personal fitness, facilitating more affordable, accessible, safer, and simpler exercise routines. This is particularly relevant for novices and individuals with disabilities. Previous research in the field is mostly dominated by the reliance on mounted sensors and a limited scope of exercises, reducing practicality for everyday use. Furthermore, existing MGAP methodologies suffer from a similar dependency on sensors and a restricted range of muscle groups, often excluding strength training exercises, which are pivotal for a comprehensive fitness regimen. Addressing these limitations, our research employs a video-based deep learning framework that encompasses a broad spectrum of exercises and muscle groups, including those vital for strength training. Utilizing the \enquote{Workout/Exercises Video} dataset, our approach integrates the X3D and SlowFast video activity recognition models in an effective way to enhance exercise classification and MGAP performance. Our findings demonstrate that this hybrid method, obtained via weighted ensemble, outperforms existing baseline models in accuracy. Pretrained models play a crucial role in enhancing overall performance, with optimal channel reduction values for the SlowFast model identified near 10. Through an ablation study that explores fine-tuning, we further elucidate the interrelation between the two tasks. Our composite model, a weighted-average ensemble of X3D and SlowFast, sets a new benchmark in both exercise classification and MGAP across all evaluated categories, offering a robust solution to the limitations of previous approaches.
\end{abstract}
\begin{document}

\flushbottom
\maketitle
%
%
\thispagestyle{empty}

\section*{Introduction}

Predicting muscle group activation allows one to improve their health and well-being. Being able to accurately identify which muscles are being activated by a movement entails more efficient methods of muscle training. If someone is shown the muscle groups that have been activated based on a certain exercise, they can modify their form to better target whichever groups they are more interested in working on at the time. For example, a person using a bench press to target their chest can modify their form to maximize the amount of chest activation.

This approach helps both people performing targeted muscle training and those new to personal fitness in achieving exercise goals in the most efficient manner \cite{van2021correcting}. Those training for muscle specialization ensure that they do not waste energy on unwanted training and those new to personal fitness can more easily determine if they are performing actions with the correct form. Optimal training leads to faster progress and achievement of benchmarks, leading to an increase in motivation. This ensures consistency with exercise, as many are motivated through results and progress \cite{teixeira2012exercise}.

In addition, usage of such technology ensures safety for those working out. Improper form can lead to unwanted muscle group activation during an exercise and prolonged repetition of such improper movements would lead to harmful consequences. For example, increased tension applied to unintended muscle groups can result in muscle injury \cite{HealthHub_2023,Mayo_Clinic_2022}. Preventing such issues is especially important for those newer to exercise, as the lack of experience means that they may develop a bad habit of performing improper form, thus increasing the risk of injury \cite{van2021correcting}.

Furthermore, muscle group activation prediction (MGAP) technology can help disabled communities with personal fitness. Physical activity is essential for a healthy life, and physical inactivity is known to cause 5.3 million deaths per year worldwide \cite{ginis2021participation}. A simple way to stay physically active is through exercise. However, many exercises are tailored for those without disabilities, making it more difficult for those with disabilities to effectively improve their health. Due to a lack of social support, people with disabilities find gyms to be intimidating and a decrease in motivation to exercise \cite{kennedy2022social}. Thus, they are $16-62\%$ less likely to achieve proper amounts of physical activity when compared to those without disabilities \cite{ginis2021participation}. The technology we present would allow them to better understand how to target different muscle groups based on whichever exercises are possible for them, and would also provide feedback on whether or not the proper movement is being performed. For example, a person with a frozen shoulder can use the technology to determine which chest exercises they can perform and ensure that the proper muscles are being utilized in their movements.

Current methods that are widely followed are limited in their practicality. One is by gaining experience and being able to self-evaluate \cite{van2021correcting}. However, this process is very time-consuming and inefficient. Another method is to hire a personal trainer \cite{waryasz2016personal}. This method is not as time-consuming but can be very costly. The US average hourly rate for a personal trainer is \$65 \cite{Gardner_2023}. In addition, the effectiveness of a personal trainer is generally dependent on cost, and thus the best options can be the highest in price \cite{Carreras_2023}.  A final method is the usage of additional sensors, such as EMG or anaerobic \cite{chiquier2023muscles,turksoy2015classification}. However, this method is difficult for many to utilize and is not practical.

In this paper, we propose a simple yet effective way of predicting muscle group activation. This is done by using video-based data of people performing exercises to train a deep learning model to classify exercises and muscle groups activated. This method is practical, as it only requires the usage of a smartphone: a common item everyone possesses. Thus, it provides the most benefit and is the most reasonable way of helping people improve their health. Moreover, the integration of exercise classification into our approach means that for common exercises, feedback can be automated. The model would identify the exercise being performed and the muscle groups being activated, allowing it to compare the ideal with the actual automatically and provide feedback.

The primary contributions of the work can be summed up as follows:
\begin{itemize}
    \item{We propose a simple and effective method for classifying exercises and predicting muscle group activation. This task is performed by combining the strengths of existing algorithms X3D and SlowFast and training the hybrid model from video data of people performing exercises. We improve the performance of this method by using a weighted-average ensemble model comprising of the two. This method is practical as its usage only requires a smartphone: a common item. 
    }
    \item{We perform comprehensive investigations that show that X3D and SlowFast achieve higher performance than all other baseline models in both exercise classification and muscle group activation prediction. In addition, our results show how the usage of a hybrid, weighted-average ensemble model combining the two further improves performance and sets a new standard for the tasks of exercise classification and muscle group activation prediction.}
\end{itemize}
Our paper is outlined as follows: 
Section 2 describes the related works in the area. Section 3 discusses our problem formulation, the data used, implementation details, and the structure of the proposed models. Section 4 analyses the model performance. Section 5 describes a discussion of the results followed by Section 6 which includes our conclusions and take-away points.

\section{Related works}
\subsection{Exercise Classification}
Prior works in exercise classification focus on two main approaches. The first uses external sensors to provide data for the model. Sensors include bands \cite{um2017exercise}, anaerobic sensors \cite{turksoy2015classification}, or IMU (inertial measurement unit) sensors \cite{preatoni2020supervised}. The data is processed through traditional machine learning models, such as a KNN. A primary drawback of this method is its practicality, as the usage of such sensors in widespread real-world applications is not pragmatic. The second approach addresses this by utilizing video-based classification and pose correction. They use video data of people performing exercises and run it through a model, such as a GCN (Graph Convolutional Network). However, these works can only account for a small number of exercises, thus reducing their effectiveness \cite{zhao20223d,rangari2022video}. In this paper, we perform video-based exercise classification for 16 different exercises. This ensures practicality and effectiveness for our approach's applications.

\subsection{Muscle Group Activation Prediction}
There are currently not enough works in the area of video-based muscle group activation prediction. Among those that are pre-existing, two approaches are present. The first approach utilizes sEMG data from sensors attached to participants. The time series data provided by the sensors is used as a label, and a temporal CNN (Convolutional Neural Network) uses video data to make predictions \cite{chiquier2023muscles}. This approach has multiple downsides as mentioned earlier. Attaching sEMG sensors is not practical for any applications. In addition, few muscle groups are used and the level of complexity for the exercises is minimal. The other approach uses video data with preassigned labels as input to a Deep Neural Network (DNN) to predict muscle group activation in each clip \cite{peng2023musclemap}. This approach is similar to what is presented in this paper. However, previous works that utilize this method are limited in the complexity of the exercises used. The authors do not include exercises for strength training, which are rather very common. To this end, in our paper, we present a method that uses the practicality of purely video-based methods to classify muscle group activation across a wide variety of exercises.

\subsection{Activity Recognition}
Since our work is closely related to activity recognition, we provide a brief review of such works. Many previous works have used videos as inputs to deep neural networks to classify actions. The DNN used can be a 2D CNN applied to single frames, a 3D CNN that accounts for temporal data, or more recent methods such as Transformers \cite{fan2021multiscale,li2022mvitv2}. One of the models we focus on in this paper, SlowFast \cite{feichtenhofer2019slowfast}, uses two 3D CNNs: one focused on the spatial dimension and one focused on the temporal dimension. The other model, X3D \cite{feichtenhofer2020x3d}, expands a 2D image classification network along spatial and temporal axes to become a 3D spatiotemporal network in such a way that optimizes model performance and efficiency at the same time.
    
\section{Problem formulation}
Our task is twofold. Firstly, we take upon the challenge of exercise classification. Prior works in this area either rely on sensors, which are impractical, or use video classification but only account for limited exercises. The goal of our version of this problem is to accurately predict exercises based on video data of them being performed. Thus, we formulate this problem to be a multi-class classification problem where a model uses spatial-temporal (video) data to accurately categorize the exercise. 

We also take upon the challenge of muscle group activation prediction (MGAP). Like exercise classification, prior works in this area use sensors for training, which is impractical, or only account for a limited number of muscle groups. Works that incorporate many muscle groups do not use exercises that need weights, which are a very common form. The goal of our version of this problem is to accurately predict muscle group activation based on a video of a person performing an exercise for various exercises and muscle groups. We formulate this problem to be a multilabel classification problem where a model uses spatial-temporal (video) data to accurately identify multiple categories of muscle groups that are simultaneously activated within one video clip. 

\section{Materials and Methods}
\subsection{Datasets}
\begin{figure}
\centering
\begin{subfigure}{.5\textwidth}
  \centering
  \includegraphics[width=.8\linewidth]{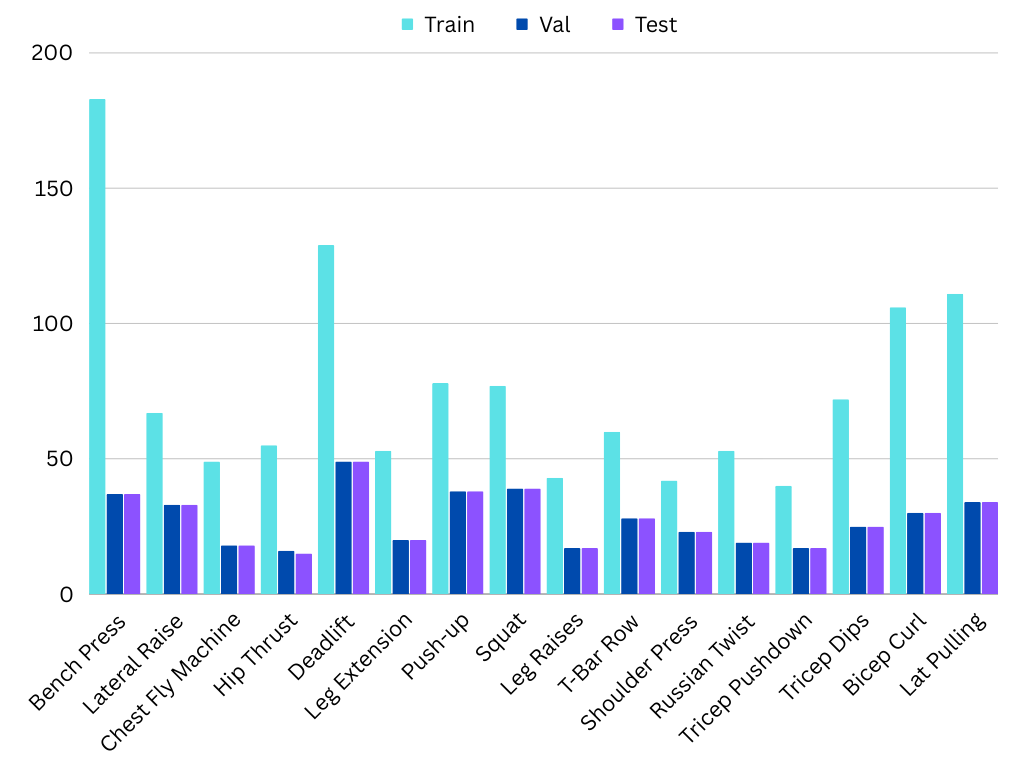}
  \caption{Exercise Classification}
  \label{fig:sub1}
\end{subfigure}%
\begin{subfigure}{.5\textwidth}
  \centering
  \includegraphics[width=.8\linewidth]{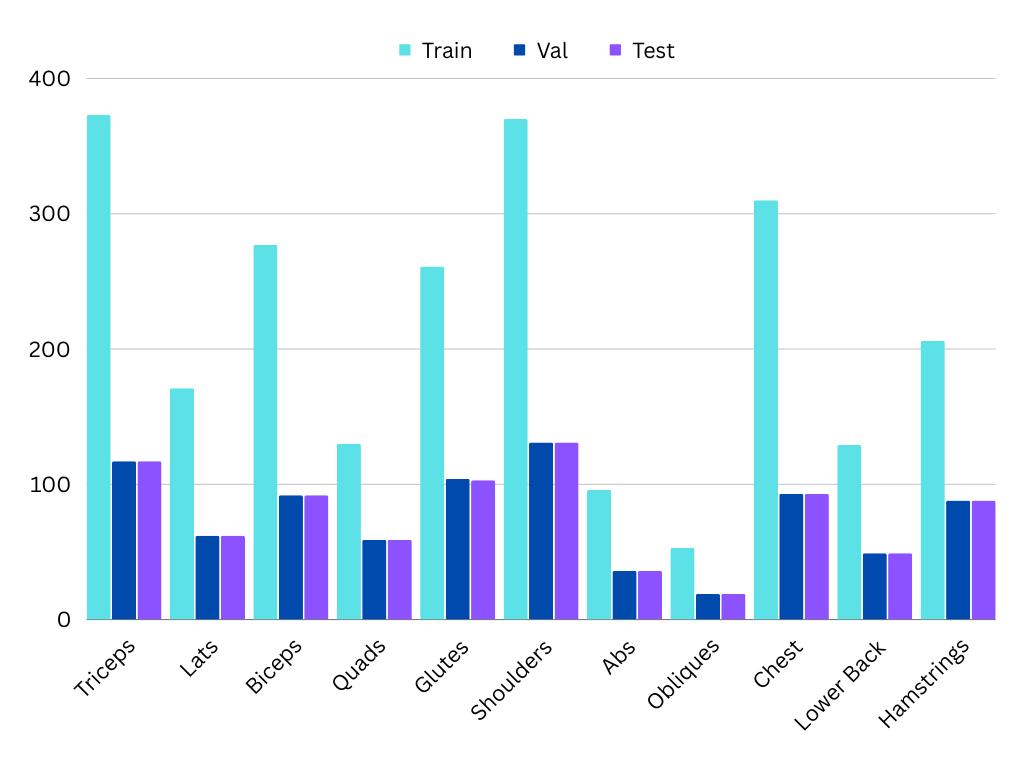}
  \caption{MGAP}
  \label{fig:sub2}
\end{subfigure}
\caption{Class Distributions in Dataset}
\label{fig:test}
\end{figure}

We use \enquote{Workout/Exercises Video} \cite{hasyim2023}, a publicly available Kaggle dataset containing videos for 22 different exercises. For the exercise classification task, we use 16 classes from this dataset. This is done by removing poor data samples and grouping similar exercises (e.g. incline, decline, and regular bench) together. We split videos to 2 seconds of length at a minimum per clip for training. For muscle-group activation prediction, we assign muscle groups to each exercise using MuscleWiki \cite{MuscleWiki} and use the same dataset. For our 16 exercise classes, we have 11 different muscle group labels in total, and the corresponding muscle group labels for each exercise can be seen in Table \ref{tab:key}. In both scenarios, there are 2113 video clips in total. 1227 are in the training split, 443 are in the validation split, and 442 are in the testing split. Figure \ref{fig:sub1} displays the class distribution for each of the splits (training, testing, validation) in exercise classification. Figure \ref{fig:sub2} displays the dataset distribution for each of the splits in MGAP.

\subsection{X3D Methodology}
Our approach uses the X3D model \cite{feichtenhofer2020x3d} developed by the Facebook AI Research lab. We use both X3D-m and X3D-s, where x3D-m is a larger model. The model is described in this section.

\subsubsection{Basis Network Instantiation} The instantiation of the basis network is called X2D, which serves as a baseline for the model before it is expanded to work with 3-dimensional data. It has a ResNet structure, similar to that of the Fast pathway (described later in \enquote{SlowFast Methodology}), with a single frame of temporal input. This is expanded to spatio-temporal network X3D utilizing the expansion factor \textit{viz}. $\gamma_T$, $\gamma_t$, $\gamma_s$, $\gamma_w$, $\gamma_b$, $\gamma_d$ . The expansion factors are discussed more in \enquote{Expansion Factors}.

The X2D model can be seen as an image classification model, as it only takes in a single frame of input. The frame is sampled with a frame rate of $1/\gamma_T$. The size of this frame can be calculated by $T \times S^2$, where $T$ denotes the temporal length and $S$ represents the dimension of a square image. In this case, the $T=1$ since only one frame is being considered. The width of individual layers is oriented similarly to the design in the Fast pathway. The width is increased by a factor of 2 after each spatial sub-sampling. In addition, the model preserves temporal input resolution throughout the network. The X2D network has a stage-level and bottleneck design. This design is similar to that of MobileNet \cite{howard2019searching} as it extends every spatial $3 \times 3$ convolution in the bottleneck block to one of $3 \times 3 \times 3$ spatiotemporal convolution.

\subsubsection{Expansion Factors}

There are 6 different expansion factors, ($\gamma_T$, $\gamma_t$, $\gamma_s$, $\gamma_w$, $\gamma_b$, $\gamma_d$), whose values are changed by 6 different expansion operations to sequentially expand a spatial X2D network into a spatiotemporal X3D network. These expansion operations consist of X-Fast, X-Temporal, X-Spatial, X-Depth, X-Width, and X-Bottleneck. X-Fast expands temporal activation size, $\gamma_t$, through increasing frame rate, $1/\gamma_T$. This increases temporal resolution while keeping clip duration constant. X-Temporal expands the temporal size, $\gamma_t$, by sampling a longer temporal clip as well as increasing frame rate, $1/\gamma_T$. This expands both temporal resolution and clip duration. X-Spatial expands spatial resolution, $\gamma_s$, by increasing the spatial sampling resolution of the input video. X-Depth expands the depth of the network by increasing the number of layers per residual stage by a factor of $\gamma_d$. X-Width uniformly expands the channel number of all the layers by a global width expansion factor of $\gamma_w$. X-Bottleneck expands the inner channel width, $\gamma_b$, of the center convolutional filter in each residual block.

\subsubsection{Progressive Expansion} Network expansion uses a simple progressive algorithm that has both forward expansion and backward contraction. X2D is used to start, with a set of the initial expansion factors $X_0 = \{\gamma_T, \gamma_t, \gamma_s, \gamma_w, \gamma_b, \gamma_d\}$. For further expansions, this can be generalized to a set $X = \{\gamma_T, \gamma_t, \gamma_s, \gamma_w, \gamma_b, \gamma_d\}$. 

For forward expansion, two functions of $X$ are used: $J(X)$ and $C(X)$. $J(X)$ measures how good the current set of expansion factors $X$ is, where higher values mean better expansion factors. This function is used to optimize model performance. $C(X)$ measures complexity from current expanding factors $X$, and is calculated as the floating point operations of the network instantiation expanded by $X$. This function is used to optimize model efficiency and keep it as lightweight as possible. Using these functions, the model expansion tries to find expansion factors $X$ with the best trade-off between $C(X)$ and $J(X)$. $X = \arg \max_{Z, C(Z)=c} J(Z)$, where $Z$ indicates the possible expansion factors to be tested and $c$ is a target complexity. Expansion is done in a progressive manner.

Backward contraction of the expansion features is used to decrease complexity and meet a desired complexity value. This is because the forward expansion is completed in discrete steps and so if the complexity target is exceeded by forward expansion, the backward contraction brings it back down to the target. This is done through a simple reduction of the last expansion till it meets the target complexity value.

\subsection{SlowFast Methodology}
Our approach uses the SlowFast \cite{feichtenhofer2019slowfast} model developed by the Facebook AI Research lab. We use both the SlowFast-R50 and SlowFast-R101 models in our approach, with the difference being the ResNet used and size of each model. The model is described in this section.

\subsubsection{Slow pathway}
The Slow pathway is a convolutional network where the key feature is a large temporal stride $\tau$. This implies that the model only processes one out of every $\tau$ frame thereby rendering the clip length to $T \times \tau$, where $T$ is the number of frames sampled. This component of the model is meant to have a high spatial resolution but a low temporal resolution. We use a $T$ value of 16 frames and a $\tau$ value of 2.

\subsubsection{Fast pathway}
The Fast pathway is another convolutional network that focuses on having a high temporal resolution. This means that the temporal stride used is much smaller for this pathway, and can be denoted as $\tau/\alpha$, where $\alpha=4$ in our case. Since the length of the clip is constant, there are always $\alpha \times T$ frames along the temporal dimension. The $\alpha$ value is a key difference between the Fast and Slow pathways.

Another key difference between the two pathways is the channel capacity. The Fast pathway has a much smaller channel capacity and can be represented as $\beta$ ($\beta < 1$) times the channel capacity of the Slow pathway. 
This means that the Fast pathway is much more lightweight, and only accounts for around $20\%$ of the total computations. In our case, $\beta = \frac{1}{8}$. This also means that the Fast pathway is weaker at representing spatial data, resulting in a stronger temporal modeling ability. 

\subsubsection{Lateral Connections}
To ensure that the temporal information learned by the Fast pathway and the spatial information from the Slow pathway do not remain isolated from each other, the two pathways are fused via lateral connections. 

A lateral connection is attached between the two pathways at every \enquote{stage}, which is right after $pool_1$, $res_1$, $res_2$, $res_3$, and $res_4$. Because of the differing temporal dimensions of the two pathways, the connections perform a transformation to match the temporal dimension. The lateral connections are unidirectional and fuse features of the Fast pathway to the Slow pathway.

At the end of the two pathways, global average pooling is performed on the outputs of each pathway. These two feature vectors are combined and then used as input for a fully connected layer that performs classification.

\subsection{Hybrid X3D-SlowFast network}
Our approach develops an X3D and SlowFast ensemble model, which improves overall performance in exercise classification and MGAP.

After evaluating the X3D and SlowFast models individually, we notice that the X3D models perform comparably to their SlowFast counterparts and even outperform them in some categories for exercise classification. We believe that the unique design of each of the networks allows it to learn certain patterns about the data that the other network cannot and that combining their results would mean a combination of information and thereby improved performance. We chose to make the model a customizable weighted average one rather than a regular average (50/50 weightage to each model) after seeing results for X3D and SlowFast in MGAP, where SlowFast consistently outperforms X3D significantly. This implies that the importance of the information learned by each model would be different, and thus their weightage should be different as well.

This hybrid model is created by taking a weighted average of the outputs of the X3D and SlowFast models. The output to our ensemble model can be represented as $\vec{V}$, where $\vec{V} = x\vec{X} + s\vec{S}$. $\vec{X}$ represents the output vector from the final layer of the X3D network before Softmax is applied and $\vec{S}$ represents the output vector from the final layer of the SlowFast network before Softmax is applied. Both $\vec{X}$ and $\vec{S}$ are $n$-vectors where $n$ is equal to the number of classes. For exercise classification $n=16$ and for MGAP $n=11$. The scalars $x$ and $s$ represent the weightage given to the X3D and SlowFast networks, respectively, and $x,s<1$ such that $x + s = 1$. The values for $x$ and $s$ that we report are (a) $x=0.75$ and $s=0.25$, (b) $x=0.5$ and $s=0.5$, and (c) $x=0.25$ and $s=0.75$.


\subsection{Training and Implementation Details}
For both exercise classification and MGAP, the videos are first normalized by subtracting the mean value of each channel from each pixel's value in that channel and then dividing that value by the standard deviation of the channel. Then, the spatial dimension of the video data is resized to $256 \times 256$. In addition, 32 frames are uniformly sub-sampled from the temporal dimension of the data. We subsample 32 frames because it is the number of frames that the pretrained SlowFast model, which we use, is designed for. For MGAP, the default label is converted to a tensor of length 11 to account for the multi-label situation. The portion of each clip that is inputted for training is 2 seconds in length and is sampled randomly from the clip. Each clip is of time $t$ seconds where $2 \le t < 4$.
Unless otherwise stated, the model is initialized with pre-trained weights on the Kinetics-400 dataset. The output layer is adjusted to the number of categories for the given task. A 16-dimensional linear classification head is used in exercise classification, and 11-dimensional classification head is used in MGAP. For the ensemble model between X3D and SlowFast, we perform a weighted average of the individual output prediction. 

We use a batch size of 8 and a learning rate of 0.0001. The code is implemented with PyTorch Lightning \cite{Falcon_PyTorch_Lightning_2019} and PyTorchVideo \cite{fan2021pytorchvideo}. The \enquote{accelerator} and \enquote{strategy} parameters are both \enquote{auto} from the PyTorch Lightning Trainer module. Early stopping for monitoring validation loss is used with a minimum charge of 0 and patience of 10. We train each model for at least 30 epochs to ensure model convergence on the V100 and A100 GPUs from Google Colab \cite{google-colab}. The code will be made publicly available upon acceptance.

\section{Experiments and Results}
In this section we present the experimental results for the two tasks, exercise classification and muscle group activation prediction, by comparing the proposed methods with baseline models and performing detailed ablation studies.

\subsection{Exercise Classification} Table \ref{tab:ec} compares the ensemble classification models with other baseline models for exercise classification. To ensure a fair comparison, all models are trained to 30 epochs. Unless \enquote{without pretraining} is specified, all models are pretrained on the Kinetics-400 dataset \cite{kay2017kinetics}. Results show that all pretrained models achieve comparable values for \enquote{Top 5 Accuracy} and \enquote{AUC}.

It can be observed that Slow-R50, which is a 3D ResNet-50 model \cite{hara2017learning}, performs the worst among the baseline models in terms of \enquote{Top 1 Accuracy}. In \enquote{Top 5 Accuracy} and \enquote{Recall}, Slow-R50 ties for last alongside I3D-R50 \cite{carreira2017quo}. I3D-R50 performed 2.75\% higher than Slow-R50 for \enquote{Top 1 Accuracy}, and performed the same in \enquote{Top 5 Accuracy} and \enquote{Recall}. However, it performed 0.12\% worse in \enquote{AUC}, 0.35\% worse in \enquote{Precision}, and 0.41\% worse in \enquote{F1 Score} when compared to Slow-R50. Slow-R50 and I3D-R50 may have been the least performing models due to them being some of the least complex tested. This is also seen in their results on Kinetics when compared to other activity recognition models \cite{feichtenhofer2020x3d}. This decreased complexity potentially meant that they were unable to learn enough information to perform the task of exercise classification effectively, and hence the lower performance. This is a probable cause as all the higher-performing models are more complex than these two.

The next two models in terms of performance are SlowFast-R50 \cite{feichtenhofer2019slowfast} and X3D-s \cite{feichtenhofer2020x3d}. SlowFast-R50 achieves an improvement of 3.31\% in accuracy with respect to I3D-R50. X3D-s improves on this by 0.45\% in the same category. Both of these models also perform very well in the \enquote{Top 5 Accuracy} category, with the highest result of 100\%. X3D-s performs worse than both Slow-R50 and I3D-R50 in the \enquote{AUC} category where it achieves a 99.29\%: the lowest performance by a pretrained model in the category. SlowFast-R50, on the other hand, improves on the previous scores by 0.2\% in the category. Both X3D-s and SlowFast-R50 perform similarly for the \enquote{Precision} category, improving on previous results by around 1.5\%. SlowFast-R50 improves in the \enquote{Recall} category by 1.41\% and in the \enquote{F1 Score} category by 2.41\%. However, it performs worse than X3D-s in these categories by 0.68\% and 0.48\%, respectively.

The next best performing model is R(2+1)D-R50 \cite{tran2018closer_r2plus1d}, which improves on X3D-s's performance in the \enquote{Top 1 Accuracy} category by 2.55\%. In the \enquote{Top 5 Accuracy} category, it performs comparably to X3D-s and SlowFast-R50. In the \enquote{AUC} category, R(2+1)D-R50 improves by only 0.02\%, performing very similar to SlowFast-R50. For the categories \enquote{Precision}, \enquote{Recall}, and \enquote{F1 Score}, R(2+1)D-R50 performs 0.46\%, 0.5\%, and 0.93\% higher than the best scores discussed thus far, respectively. A reason for R(2+1)D-R50's high performance could be due to its similarities to the SlowFast model, which is the highest-performing model overall. Both R(2+1)D and SlowFast models consider the spatial and temporal dimensions separately. Thus, this similar approach to video classification could be especially effective for our task.

Overall, the best performing baseline models for this task are X3D-m \cite{feichtenhofer2020x3d} and SlowFast-R101 \cite{feichtenhofer2019slowfast}. X3D-m improved on R(2+1)D-R50 by 0.32\% in the \enquote{Top 1 Accuracy} category. SlowFast-R101 further improved by another 0.56\% in the category. X3D-m performed better than SlowFast-R101 in the \enquote{Top 5 Accuracy} category, but the two values are comparable. In the \enquote{AUC} category, X3D-m improved on R(2+1)D-R50 by 0.17\%, and SlowFast-R101 performed similarly at 99.85\% (0.04\% less). In terms of \enquote{Precision}, X3D-m improved on R(2+1)D-R50 by 0.43\%, and SlowFast-R101 performed similarly at 98.86\% (0.08\% less). In terms of \enquote{Recall}, X3D-m improved on R(2+1)D-R50's performance by 0.13\%, and SlowFast-R101 further improved on it by another 0.23\%. In terms of \enquote{F1 Score}, X3D-m improved on R(2+1)D by 0.7\% and performed similarly to SlowFast-R101 (which was only 0.05\% higher).

Table \ref{tab:ec} also shows the results for the exercise classification ensemble model as discussed in $\S IV.E$. The last three rows show the different types of ensemble models. The number that indicates it is the ratio of percentages for the weighted average. The first number shows the percentage for X3D-m and the second corresponds to SlowFast-R101. This means that the smaller the value of the fraction the more weight given to the SlowFast model. We choose to use X3D-m and SlowFast-R101 to form the ensemble model because they are the best-performing baseline models in exercise classification. A key observation to note is that, overall, the ensemble models are some of the best-performing models. They outperform all baseline models (besides X3D-m and SlowFast-R101) in all categories. X3D-m and SlowFast-R101 outperform or perform comparably in almost all categories.

The \enquote{25/75} model performs either better or comparably to all the other ensemble models. All ensemble models perform the same in \enquote{Top 5 Accuracy}, and there is little to no difference in performance for \enquote{AUC} between the \enquote{50/50} model and the \enquote{25/75} model. Between the \enquote{50/50} and \enquote{75/25} models in the \enquote{AUC} category, the \enquote{50/50} model only performs slightly better (by 0.33$\%$).  In the \enquote{Top 1 Accuracy} category, the \enquote{25/75} model outperforms \enquote{75/25} by $2.27\%$ and \enquote{50/50} by $0.57\%$. In terms of \enquote{Precision}, the \enquote{25/75} model outperforms \enquote{75/25} by $0.82\%$ and \enquote{50/50} by $0.34\%$. In terms of \enquote{Recall}, the \enquote{25/75} model outperforms \enquote{75/25} by $0.66\%$ and \enquote{50/50} by $0.24\%$. In the \enquote{F1 Score} category, the \enquote{25/75} model outperforms \enquote{75/25} by $1.46\%$ and \enquote{50/50} by $0.69\%$. When the results of the \enquote{25/75} model are compared to the best results in each category for X3D-m and SlowFast-R101 (e.g. whichever value is bigger among the two), it can be seen to perform either comparably or better. It performs similarly in the categories \enquote{Top 5 Accuracy} and \enquote{AUC}. For the other categories, it performs better by $1.23\%$ in \enquote{Top 1 Accuracy}, $0.81\%$ in \enquote{Precision}, $0.87\%$ in \enquote{Recall}, and $1.5\%$ in \enquote{F1 Score}. These results show that overall the features extracted by SlowFast are more valuable than those by X3D, but the X3D knowledge is still necessary for an improvement in performance.

A significant aspect of the experiments is the importance of pretraining. When comparing the SlowFast models that are not trained on Kinetics-400 and those that are, one can see an astonishing improvement that comes from pretraining. Pretraining increases performance in \enquote{Top 1 Accuracy} by around 40\%. For the \enquote{Top 5 Accuracy} category, pretraining improves performance by around 25\%. For the metrics \enquote{AUC}, \enquote{Precision}, and \enquote{Recall}, pretraining improves performance by around 15\%. For the \enquote{F1 Score} category, pretraining improves performance by around 30\%. This can also be seen when comparing the pretrained and non-pretrained X3D models, where the improvement is much more substantial. Both models improve by almost 60\% in the \enquote{Top 1 Accuracy} category and by 50\% in the \enquote{Top 5 Accuracy} category. In terms of \enquote{AUC}, \enquote{Precision}, \enquote{Recall}, and \enquote{F1 Score}, the pretrained X3D models improve on the non-pretrained ones by around 30\%, 20\%, 20\%, and 30\%, respectively. Thus, we observe that using models that are pretrained on a similar task (both exercise classification and Kinetics are activity recognition) can prove to be very beneficial in increasing model performance even though the specific downstream application is unrelated.

\subsection{Muscle Group Activation Prediction}  Table \ref{tab:mgap} compares the SlowFast models with other baseline models for the MGAP task. To ensure a fair comparison, all models are trained for 30 epochs. Also, unless \enquote{without pretraining} is specified, all models are pretrained on the Kinetics-400 dataset. 

As in the exercise classification case, Slow-R50 and I3D-R50 perform significantly worse than X3D and SlowFast. Despite performing well in the exercise classification task, R(2+1)D-R50 achieves poor performance in terms of precision, recall, and F1 score for MGAP.
Despite the similarities in the approaches of R(2+1)D and SlowFast, this difference in performance potentially means that the intricacies behind the SlowFast approach are more influential for the MGAP task. 

The next highest-performing models are the X3D models. In all categories, except \enquote{Recall} (where I3D-R50 holds the highest score thus far), X3D-s improves on performances over the baselines. It improves in \enquote{Accuracy} by 0.68\%, in \enquote{AUC} by 0.73\%, in \enquote{Precision} by 0.92\%, and in \enquote{F1 Score} by 1.45\%. X3D-m improves further on these scores further by 0.97\%, 3.01\%, 3.5\%, and 2.77\%, respectively. In the \enquote{Recall} category, X3D-m improves on X3D-s by 2.39\%. Thus, it is observed that a larger X3D model results in a significant performance improvement overall. 

For the pretrained SlowFast models, we observe that SlowFast-R50 performs higher in the \enquote{Accuracy} category than its -R101 counterpart by 0.85\%. SlowFast-R50 also performs higher than SlowFast-R101 in the \enquote{AUC} category by 1.11\%. However, for the metrics \enquote{Precision}, \enquote{Recall}, and \enquote{F1 Score}, SlowFast-R101 performs higher than SlowFast-R50 by 0.23\%, 1.77\%, and 0.68\%, respectively. However, unlike the X3D models, the additional number of trainable parameters of the -R101 model did not cause a substantial improvement for the SlowFast models.

Table \ref{tab:mgap} also shows the results for the MGAP ensemble model. The MGAP ensemble model is made up of the X3D-m and SlowFast-R50 models, as those are the best X3D and SlowFast versions for MGAP. When comparing the SlowFast-R50 and X3D-m models, one can see that SlowFast-R50 outperforms X3D-m in every metric. This is unlike the results for exercise classification, in which the X3D model outperformed SlowFast in many evaluation metrics. As with exercise classification, a key observation is that, overall, the ensemble models outperform most baselines in all categories.

It can be seen that the \enquote{25/75} model is the best of the ensemble models in all categories. The \enquote{25/75} ensemble model performs comparably to \enquote{50/50} in the categories \enquote{AUC} and \enquote{Recall}. In these categories, it performs better than \enquote{75/25} by 0.88\% and 3.02\%, respectively. In terms of \enquote{Accuracy}, the \enquote{25/75} model performs better than the \enquote{75/25} model by 2.97\% and the \enquote{50/50} model by 1.33\%. In \enquote{Precision}, the \enquote{25/75} model performs better than the \enquote{75/25} model by 7.46\% and the \enquote{50/50} model by 1.97\%. In the \enquote{F1 Score} category, the \enquote{25/75} model performs better than the \enquote{75/25} model by 5.93\% and the \enquote{50/50} model by 1.31\%. When the results of the \enquote{25/75} model are compared to the results of SlowFast-R50 (which is the best non-ensemble model in Table \ref{tab:mgap}), it can be seen to perform either comparably or better. It performs comparably in the categories \enquote{Accuracy} and \enquote{AUC}, and performs better in the categories \enquote{Precision}, \enquote{Recall}, and \enquote{F1 Score} by 6.68\%, 7.72\%, and 7\%, respectively. It is interesting to note that the models that have a majority X3D percentage (\enquote{75/25} and \enquote{50/50}) perform similarly to the baseline SlowFast-R50 model. The \enquote{75/25} model performs slightly better in the \enquote{Recall} category, comparably in the \enquote{F1 Score} category, and worse in all others. The \enquote{50/50} model, which has an increased SlowFast percentage, performs worse in \enquote{Accuracy} and \enquote{AUC}, but better in \enquote{Precision}, \enquote{Recall}, and \enquote{F1 Score}. Similar to the exercise classification task, these results show that though the information from SlowFast is much more valuable in MGAP, some information from X3D is still necessary to improve performance.

The non-pretrained SlowFast models still perform surprisingly well, which shows their capability to learn the relevant features from scratch quickly. For the categories \enquote{Accuracy} and \enquote{AUC}, they outperform Slow-R50 and I3D-R50, both of which are pretrained on the Kinetics dataset. For the categories \enquote{Precision}, \enquote{Recall}, and \enquote{F1 Score}, the non-pretrained models outperform Slow-R50, I3D-R50, R(2+1)D-R50, X3D-s, and X3D-m, all of which are pretrained. However, the non-pretrained versions of the SlowFast models cannot surpass their pretrained counterparts in any category, similar to the results for exercise classification. In the \enquote{Accuracy}, \enquote{AUC}, and \enquote{Recall} category, pertaining results in improvements of around 20\%. In the \enquote{Precision} and \enquote{F1 Score} categories, there are improvements of around 30\%. The results of the non-pretrained X3D models show an even greater significance to using pretrained models. They are only able to outperform the \enquote{Slow-R50} and \enquote{I3D-R50} models, but perform worse in the \enquote{Recall} category. All other pretrained models outperform them in terms of every metric. When compared to their pretrained counterparts, they perform worse in \enquote{Accuracy} by around 6\% and worse by around 25\% in \enquote{AUC}. In the categories \enquote{Precision}, \enquote{Recall}, and \enquote{F1 Score}, the non-pretrained X3D models perform worse than the pretrained X3D models by around 15\%, 8\%, and 10\%, respectively. Overall, these results show the substantial benefit that occurs from using pretrained models.

\subsection{Ablation Studies}
In this section, we investigate the variation of channel reduction ratios as well as different finetuning strategies for MGAP task.

\subsubsection{Channel reduction ratios (Beta Value)} Table \ref{tab:beta} shows an ablation study involving the channel reduction ratios for SlowFast-R50 models. None of the models here are pretrained and are trained to 30 epochs. The channel reduction ratio corresponds to $1/\beta$ (inverse beta), which is previously described in the \enquote{Methodology} section. It can be seen that the best-performing channel reduction ratios are either minimal (such as 2) or at values close to 10. An inverse beta value of 8 performs highest in the \enquote{Top 1 Accuracy} category, with a performance of 51.52\%. The next highest is a value of 10, which performs only 1\% lower. A beta value of 8 also performs best in \enquote{Recall} but is only 0.43\% higher than that of 10. An inverse $\beta$ value of 10 performs the best for the categories \enquote{Top 5 Accuracy}, \enquote{AUC}, and \enquote{F1 Score}. However, it performs poorly in terms of \enquote{Precision}. Inverse beta value of 2 performs the best in terms of precision which is 1.43\% higher than that of 10. Nevertheless, this analysis shows that for the exercise classification problem we face, a channel reduction ratio of 10 is most optimal for a majority of the evaluation categories. 

\subsubsection{Transfer Learning} Table \ref{tab:ft} shows an ablation study of fine-tuned versions of the X3D and SlowFast models. We aim to explore the importance of the exercise classification task in relation to the downstream MGAP task and hence the influence of transfer learning in this context. Therefore, we pretrain the X3D and SlowFast models on the exercise classification task and then finetune it on the MGAP task. The models are trained for 50 epochs for exercise classification and the best versions are used. They are fine-tuned for 30 epochs for MGAP and then compared with the Kinetics-400 pretrained models in Table \ref{tab:ft}. 

When comparing the X3D-s models, one can see that there is a significant improvement in the \enquote{Accuracy} category, where the fine-tuned X3D-s model improves on X3D-s by 2.33\%. In fact, out of all the models displayed in this table, X3D-s performs the best in the \enquote{Accuracy} category (though the two X3D-m models show similar performance). The fine-tuned X3D-s model does not improve in the \enquote{AUC} category, where it performs 2.23\% less than the Kinetics pretrained version. However, it performs higher in the categories \enquote{Precision}, \enquote{Recall}, and \enquote{F1 Score} by 6.56\%, 3.78\%, 4.65\%, respectively. When comparing the X3D-m models, we can see that there is little improvement overall. The only category in which it improves is the \enquote{AUC} category, in which it does so by 1.63\%. In the \enquote{Accuracy} category, it performs comparably with only a 0.43\% decrease. In the categories \enquote{Precision}, \enquote{Recall}, and \enquote{F1 Score}, the fine-tuned X3D-m model performs worse by 10.04\%, 6.33\%, and 7.67\%, respectively. These results show that pre-training the model on the exercise classification task does not improve the performance of the model significantly.

When comparing the SlowFast-R50 models, it can be seen that there is a similar performance across all the categories. We observe that the fine-tuned model performs higher by 0.45\% in the \enquote{Accuracy} category and higher by 1.33\% in the \enquote{Precision} category. However, it performs worse by 0.2\%, 1.9\%, and 0.45\% in the categories \enquote{AUC}, \enquote{Recall}, and \enquote{F1 Score}. Despite the lower performance in some categories, the results are comparable and the model performances are quite similar. When comparing the SlowFast-R101 model to its fine-tuned counterpart, we can see that there is no improvement from pretraining on the exercise classification task. The fine-tuned model performs worse by 0.9\%, 1.65\%, 5.78\%, 12.51\%, and 9.13\% in the categories \enquote{Accuracy}, \enquote{AUC}, \enquote{Precision}, \enquote{Recall}, and \enquote{F1 Score}. 

Table \ref{tab:ft} also shows the ablation results for transfer learning with the MGAP ensemble models. It is comprised of fine-tuned X3D-m and SlowFast-R50 models that were initially pretrained on the exercise classification task. The last three rows show the different ensemble models indicated by the ratio of percentages for the weighted average. The first number shows the percentage for fine-tuned X3D-m and the second corresponds to fine-tuned SlowFast-R50. 

When comparing the fine-tuned SlowFast-R50 and fine-tuned X3D-m models, we can see that fine-tuned SlowFast-R50 outperforms fine-tuned X3D-m in every category. This is similar to the results for exercise classification, in which the X3D model outperformed SlowFast in many categories. 
The \enquote{25/75} ensemble model performs comparably to the \enquote{50/50} model in \enquote{Recall} and \enquote{F1 Score}. In these categories, the \enquote{25/75} model improves on the \enquote{75/25} model by 1.17\% and 1.52\%, respectively. In terms of \enquote{Accuracy}, the \enquote{25/75} model outperforms \enquote{75/25} by 0.53\% and \enquote{50/50} by 0.26\%. In the \enquote{AUC} category, the \enquote{25/75} model outperforms \enquote{75/25} by 0.64\% and \enquote{50/50} by 0.81\%, where as in the \enquote{Precision} category, the \enquote{25/75} model outperforms \enquote{75/25} by 1.89\% and \enquote{50/50} by 0.55\%. When the results of the \enquote{25/75} model are compared to the results of fine-tuned SlowFast-R50 (which is the best performing non-ensemble model in Table \ref{tab:ft}), it can be seen to perform better in every single category. In the \enquote{Accuracy} and \enquote{AUC} categories, the improvement is not very substantial with only 0.22\% and 0.16\% increase in the respective categories. However, there is much more significant improvement for the other categories. In the metrics \enquote{Precision}, \enquote{Recall}, and \enquote{F1 Score}, there are improvements of 5.61\%, 4.39\%, and 5.1\%, respectively. These results show that the information that SlowFast-R50 learned from exercise classification is much more useful than what X3D-m learned, but both models contribute towards an increase in performance. They also highlight the connectivity between exercise classification and muscle group activation prediction, as the \enquote{25/75} model (which technically only consists of information learned through exercise classification) ended up performing the best in MGAP overall.

\section{Discussion}
From the results presented, we show the abilities of X3D and SlowFast models to perform well on the exercise classification and muscle group activation prediction tasks. These models can perform these tasks to a degree of usefulness that has applications in the real world to make personal fitness easier and cheaper. In addition, it is shown that the SlowFast models can be seen to perform the best when compared to other baseline models, with the X3D models coming second in performance. Among these comparisons, the high performance of the R(2+1)D is slightly surprising. The high performance of the R(2+1)D model could be due to its similarities to the SlowFast models as it also considers spatial and temporal data differently. However, it is more surprising how this gap is much larger for the MGAP task than for exercise classification, even though an ablation study shows how connected the information learned is for both. The poor performance of R(2+1)D in this case is potentially due to the fact that the MGAP task requires stronger spatiotemporal feature extraction ability than that present in R(2+1)D models. Another interesting result is the difference in the relationship between the SlowFast model complexity and its performance. For the X3D models, in both exercise classification and MGAP the model with greater complexity performs better. However, for SlowFast, it is not as consistent. For exercise classification, larger model is superior in performance. However, for the MGAP task both -R50 and -R101 models have similar performances, with each having a higher performance in different categories. Our results also show the large benefit in performance that comes from using pretrained models, more specifically from datasets that are similar to the task at hand. For both X3D and SlowFast, the improvement that is seen when a non-pretrained and pretrained model is compared is immense, with the pretrained model performing much better. However, something unexpected is the higher performance of non-pretrained SlowFast and X3D models when compared to the Slow-R50 and I3D-R50 pretrained baseline models for the MGAP task. Though surprising, this result enforces the fact that X3D and SlowFast are the best-performing models for the task, as even their non-pretrained versions can outperform the pretrained versions of other baseline models. Our results also show how the channel reduction ratio parameter affects the performance of the SlowFast models. It can be seen that even with a small inverse $\beta$ value, such as 2, the model performs reasonably well. 
However, a higher performance can be observed with values close to 10. We noted that this is the best ratio at which the Fast pathway understands temporal data and allows the Slow pathway to focus on spatial data, thus achieving the goals behind the model. A final observation from our results is the improvement of ensemble models. For both exercise classification and MGAP, it can be seen that the use of a weighted-average ensemble model comprised of the best-performing X3D and SlowFast models of that task improves on the previously best overall results for that task The best-performing weighting for the ensemble model always had a 25\% weight on the X3D model and a 75\% weight on the SlowFast model. This demonstrates that the information learned by SlowFast is more valuable than what X3D has learned. 

\section{Conclusion}
In this paper, we investigate several video-based classification algorithms and present an effective approach to tackle the tasks of exercise classification and muscle group activation prediction. We contribute by carefully designing a hybrid methodology, that includes the X3D and SlowFast models as components of an ensemble model. 

Our method is practical for real-life applications since it is simple (only requires a smartphone), effective (performs well as shown in the experiments), and adaptive (considers a wide variety of exercises and muscle groups). Our method surpasses other baseline models for both tasks, and there is a significantly higher performance for the MGAP task when using our methodology. Our studies also show the importance of utilizing pretrained models, the optimal channel reduction ratios for SlowFast, and the relationship between the information learned across the exercise classification and MGAP tasks. We propose the use of a weighted-average ensemble model, which further improves performance in all categories and sets a new standard for exercise classification and muscle group activation prediction. 

\bibliography{sample}

\section*{Author contributions statement}
M.P. conceived and conducted the experiments. M.P., with the aid of P.S., analyzed the results. All authors reviewed the manuscript. 

\section{Data availability statement} All data is publicly available and sourced from \enquote{Workout/Exercises Video} on Kaggle (\url{https://www.kaggle.com/datasets/hasyimabdillah/workoutfitness-video}) \cite{hasyim2023}.

\section{Code availability statement} The source code is available from our Github code repository: https://github.com/ManvikPasula/ExerciseClassification-MGAP.

\section*{Additional information}

\textbf{Competing interests} The authors have no competing interests, neither financial nor non-financial.

\section{Figures}

\section{Tables}

\begin{table}[!htp]
\centering
    \caption{Exercises and Corresponding Activated Muscle Groups}\label{tab:key}
    \begin{tabular}{lrr}\toprule
        \textbf{Exercise} &\textbf{Activated Muscle Groups} \\
        Tricep Pushdown &triceps \\
        Tricep Dips &triceps \\
        T-bar Row &lats, biceps \\
        Squat &quads, glutes, hamstrings \\
        Shoulder Press &shoulders \\
        Russian Twist &abs, obliques \\
        Push-up &chest, triceps, shoulders \\
        Leg raises &abs \\
        Leg extension &quads \\
        Lateral raise &shoulders \\
        Lat Pulldown &lats, biceps \\
        Hip Thrust &glutes \\
        Deadlift &glutes, lower back, hamstrings \\
        Chest Fly &chest \\
        Bicep Curl &bicep \\
        Bench Press &chest, shoulders, triceps \\
        \bottomrule
    \end{tabular}
\end{table}

\begin{table*}[!ht]
    \centering
    \caption{Comparison with baseline and non-pretrained models for exercise classification. In the `Model' column, for the last three rows, the value on the left of the ratio corresponds to the weighted average percentage for X3D-m and the value on the right corresponds to SlowFast-R101 for the ensemble model of X3D-m and SlowFast-R101. {R} implies {ResNet}, {w/o PT} implies {without pretraining}, {Prec} implies {Precision}.}\label{tab:ec}
    \vspace{2mm}
    \begin{tabular}{lrrrrrrr}\toprule
        \textbf{Model Name} & \textbf{Top 1} (\%) &\textbf{ Top 5 }(\%) & \textbf{AUC} (\%) & \textbf{Prec} (\%) & \textbf{Recall} (\%) & \textbf{F1} (\%)\\ \hline
                X3D-s w/o PT & 33.25 & 53.75 & 72.18 & 83.93 & 78.14 & 66.38 \\
        X3D-m w/o PT & 36.25 & 63.50 & 74.01 & 87.97 & 78.08 & 69.37 \\
        SlowFast-R50 w/o PT & 51.52 & 73.96 & 85.43 & 83.88 & 82.86 & 69.99 \\  
        SlowFast-R101 w/o PT & 58.71 & 80.21 & 87.99 & 84.56 & 84.67 & 72.44 \\ \hline
        Slow-R50 & 87.50 & 99.50 & 99.50 & 96.67 & 96.15 & 93.55 \\ 
        I3D-R50 & 90.25 & 99.50 & 99.38 & 96.32 & 96.15 & 93.14 \\ 
        R(2+1)D-R50 & 96.56 & 99.66 & 99.72 & 98.51 & 98.74 & 97.37 \\ \hline

        X3D-s & 94.01 & 100.00 & 99.29 & 98.05 & 98.24 & 96.44 \\ 
        X3D-m & 96.88 & 100.00 & 99.89 & 98.94 & 98.87 & 98.07 \\ 
        SlowFast-R50 & 93.56 & 100.00 & 99.70 & 98.01 & 97.56 & 95.96 \\ 
        SlowFast-R101 & 97.44 & 99.72 & 99.85 & 98.86 & 99.10 & 98.12 \\ \hline

        75/25 &96.88 &99.72 &99.64 &98.93 &99.08 &98.11 \\
        50/50 &98.58 &99.72 &99.97 &99.36 &99.46 &98.88 \\
        25/75 &99.15 &99.72 &99.94 &99.75 &99.74 &99.57 \\        
        \bottomrule
    \end{tabular}
\end{table*}

\begin{table*}[!htp]
    \centering
    \caption{Comparison with baseline and non-pretrained models for MGAP. In the 'Model' column for the last three rows, the value on the left of the ratio corresponds to the weighted average percentage for X3D-m and the value on the right corresponds to SlowFast-R50 for the ensemble model of X3D-m and SlowFast-R50. {R} implies {ResNet}, {w/o PT} implies {without pretraining}, {Prec} implies {Precision}.}\label{tab:mgap}
    \vspace{2mm}
    \begin{tabular}{lrrrrrr}\toprule
        \textbf{Model} &\textbf{Accuracy (\%)} &\textbf{AUC (\%)} &\textbf{Prec (\%)} &\textbf{Recall (\%)} &\textbf{F1 (\%)} \\\midrule
        Slow-R50 &74.01 &56.53 &4.92 &14.96 &6.91 \\
        I3D-R50 &44.44 &48.37 &7.98 &34.28 &12.30 \\
        R(2+1)D-R50 &86.72 &91.52 &26.85 &20.12 &21.87 \\ \hline
        X3D-s w/o PT &83.82 &70.81 &16.26 &14.10 &14.05 \\
        X3D-m w/o PT &82.82 &71.19 &15.57 &13.38 &13.02 \\
        SlowFast-R50 w/o PT &75.97 &79.15 &30.90 &42.29 &33.44 \\
        SlowFast-R101 w/o PT &80.14 &84.47 &32.15 &41.52 &33.56 \\ \hline
        X3D-s &87.40 &92.25 &27.77 &21.63 &23.32 \\
        X3D-m &88.37 &95.26 &31.27 &24.02 &26.09 \\
        SlowFast-R50 &97.11 &99.43 &58.19 &61.33 &59.05 \\
        SlowFast-R101 &96.26 &98.32 &58.42 &63.10 &59.73 \\ \hline

        75/25 &94.46 &98.22 &57.47 &64.11 &59.35 \\
        50/50 &96.10 &99.05 &62.96 &67.53 &63.97 \\
        25/75 &97.43 &99.10 &64.93 &67.13 &65.28 \\
        \bottomrule
    \end{tabular}
\end{table*}

\begin{table*}[!htb]\centering
    \caption{Ablation study for testing different SlowFast channel reduction values on exercise classification. This number corresponds to the inverse of the $\beta$ value. {Prec} implies {Precision}.}\label{tab:beta}
    \vspace{2mm}
    \begin{tabular}{m{2.3cm}lrrrrrr}\toprule
        \textbf{Inverse Beta} &\textbf{Top 1 (\%)} &\textbf{Top 5 (\%)} &\textbf{AUC (\%)} &\textbf{Prec (\%)} &\textbf{Recall (\%)} &\textbf{F1 (\%)} \\\midrule
        2 &49.53 &73.30 &86.22 &84.58 &82.10 &70.16 \\
        4 &43.37 &74.43 &83.62 &82.68 &78.98 &65.57 \\
        8 &51.52 &73.96 &85.43 &83.88 &82.86 &69.99 \\
        10 &50.52 &82.29 &88.44 &83.15 &82.43 &70.98 \\
        16 &47.35 &79.92 &86.51 &83.93 &80.37 &68.43 \\
    \bottomrule
    \end{tabular}
\end{table*}

\begin{table*}[!htp]\centering
    \caption{Ablation for testing a fine-tuned version of the X3D and SlowFast models that adapt the exercise classification model for the MGAP task. In the 'Model' column for the last three rows, the value on the left of the ratio corresponds to the weighted average percentage for Fine-Tuned X3D-m and the value on the right corresponds to Fine-Tuned SlowFast-R50 for the ensemble model. {Prec} implies {Precision} and {FT} implies {Fine-Tuned}.}\label{tab:ft}
        \vspace{2mm}
        \begin{tabular}{lrrrrrr}\toprule
        \textbf{Model} &\textbf{Accuracy (\%)} &\textbf{AUC (\%)} &\textbf{Prec (\%)} &\textbf{Recall (\%)} &\textbf{F1 (\%)} \\\midrule
        X3D-s &87.40 &92.25 &27.77 &21.63 &23.32 \\
        X3D-m &89.45 &96.63 &37.40 &27.25 &30.32 \\
        FT X3D-s &89.73 &90.02 &34.33 &25.41 &27.97 \\
        FT X3D-m &89.02 &98.28 &27.36 &20.92 &22.65 \\ \hline

        SlowFast-R50 &97.37 &99.32 &58.25 &59.41 &58.28 \\
        SlowFast-R101 &96.26 &98.32 &58.42 &63.10 &59.73 \\
        FT SlowFast-R50 &97.82 &99.12 &59.58 &57.51 &57.83 \\
        FT SlowFast-R101 &95.36 &96.68 &52.64 &50.59 &50.61 \\ \hline

        75/25 &97.50 &98.63 &63.29 &60.72 &61.41 \\
        50/50 &97.78 &98.46 &64.63 &62.15 &62.91 \\
        25/75 &98.04 &99.27 &65.19 &61.90 &62.93 \\
        \bottomrule
    \end{tabular}
\end{table*}

\end{document}